# BranchPoseNet: Characterizing tree branching with a deep learning-based pose estimation approach

Stefano Puliti*, Carolin Fischer, Rasmus Astrup

Norwegian Institute of Bioeconomy Research (NIBIO), Høgskoleveien 8, 1430 Ås, Norway

* corresponding author (stefano.puliti@nibio.no)

## Abstract

This paper presents an automated pipeline for detecting tree whorls in proximally laser scanning data using a pose-estimation deep learning model. Accurate whorl detection provides valuable insights into tree growth patterns, wood quality, and offers potential for use as a biometric marker to track trees throughout the forestry value chain. The workflow processes point cloud data to create sectional images, which are subsequently used to identify keypoints representing tree whorls and branches along the stem. The method was tested on a dataset of destructively sampled individual trees, where the whorls were located along the stems of felled trees. The results demonstrated strong potential, with accurate identification of tree whorls and precise calculation of key structural metrics, unlocking new insights and deeper levels of information from individual tree point clouds.

# 1. Introduction

Understanding and quantifying individual tree structure, with regards to crown and branching patterns, is an important and yet underexplored topic with relevance for many aspects of forest dynamics, wood quality, habitat and stand structure. Due to the difficulty in measuring tree crown variables from the ground, branching information has not previously been available beyond small samples of destructively sampled trees. High resolution laser scanning data shows great potential for capturing branching patterns. Previous research has found branching information from laser scanning useful for various applications, including: branch architecture quantification (Wilkes et al. 2021), tree species classification (Puliti et al. 2024; Terryn et al. 2020), tree growth characterization (Ahmed et al. 2024; Puliti et al. 2023a), wood quality quantification (Cattaneo et al. 2024; Pyörälä et al. 2019), and quantifying drought impacts (Kröner et al. 2024). Traditional methods for analyzing tree structures from LiDAR data often involve manual or semi-automated processes.

In recent years, deep learning approaches have shown their potential in streamlining the automatic detection of tree architectural features. Puliti at al. (2023) developed a method relying on a whorl bounding-box detection and found this method to produce accurate individual tree height growth curves useful to derive site productivity information, a key variable for forest management. One limitation of bounding-box detection is its simplistic approach to capturing tree branching patterns, missing the biological logic behind tree structures such like branch arrangement along the stem.

To address this, we transitioned to a pose estimation model, which predicts the positions of keypoints to map the structure of objects—in this case, tree branches and whorls. Pose estimation models are a powerful alternative for detecting structural features like tree whorls and branches, as they focus on identifying key points in an object's structure rather than simply drawing bounding boxes around areas of interest. These models, often powered by convolutional neural networks (CNNs), predict the location of specific keypoints, allowing them to capture the spatial relationships between different parts of a tree, such as the arrangement of branches along the stem. This method, which is commonly used in human pose estimation (Andriluka et al. 2014), leverages the existence of pre-defined fixed data structures to improve detection and offers a more detailed and biologically relevant model for tree architecture characterization.

This study introduces a fully automated workflow that uses individual tree points and a YOLOv8 pose estimation model to characterize tree whorls and their geometry. The proposed method builds conceptually on the previous work by Puliti et al. (2023) and aims at improving it in the three following ways

- **Fixed object geometry**: by relying on a pose estimation model, this new approach introduces knowledge on the biological tree structure to the object to be detected: three keypoints always in the same order and connected by a logic) and thus can rely on a larger pool of visual clues (i.e. branches) to allow the detection of individual whorls
- **Platform/sensor agnostic**: with nearly a tenfold increase in training data we boosted both the amount and variation in the data including data from multiple tree species (i.e, *Picea abies*, *Pinus sylvestris*, *Pseudotsuga menziesii*) and captured using terrestrial, mobile, and uncrewed laser scanning (TLS, MLS, and ULS).
- **Capturing whorl geometries:** Thanks to the fixed geometrical structure inbuild in the model it allows the extraction of several geometrical whorls properties (i.e. branch insertion angles and branch lengths) that are useful additional structural parameters.

# 2. Materials

## 2.1 Whorl pose estimation data

The data for developing the whorl pose detection model consisted of images of projected point cloud profiles (see m section 3.1 and figure 1c) from individual tree point cloud from terrestrial (TLS), mobile (MLS), and unmanned laser scanning (ULS) from the FOR-instance (Puliti et al. 2023b) and FOR-species (Puliti et al. 2024) datasets. From all the generated images we purposively selected 1,067 of them to ensure covering a broad range of tree developmental stages and variation in branch architectures across different parts of the tree from images from the bottom to the top of the trees. These images were labeled by a team of four annotators based on the defined pose data structure (see Figure 1a). In total 12,870 whorls and 25,740 branches were manually labelled.

## 2.2 Test data

To evaluate the developed whole-pose detection model in a forest-related downstream application we applied it to the same destructively trees as those used by Puliti et al. (2023a) for which the distance between the tree base and the whorls was manually measured on felled trees. The model's predictions were then evaluated according to the their ability to correctly predict the presence of a whorl (F1-score) and for the accuracy (RMSE; m) in the measurement of the intermodal distance.

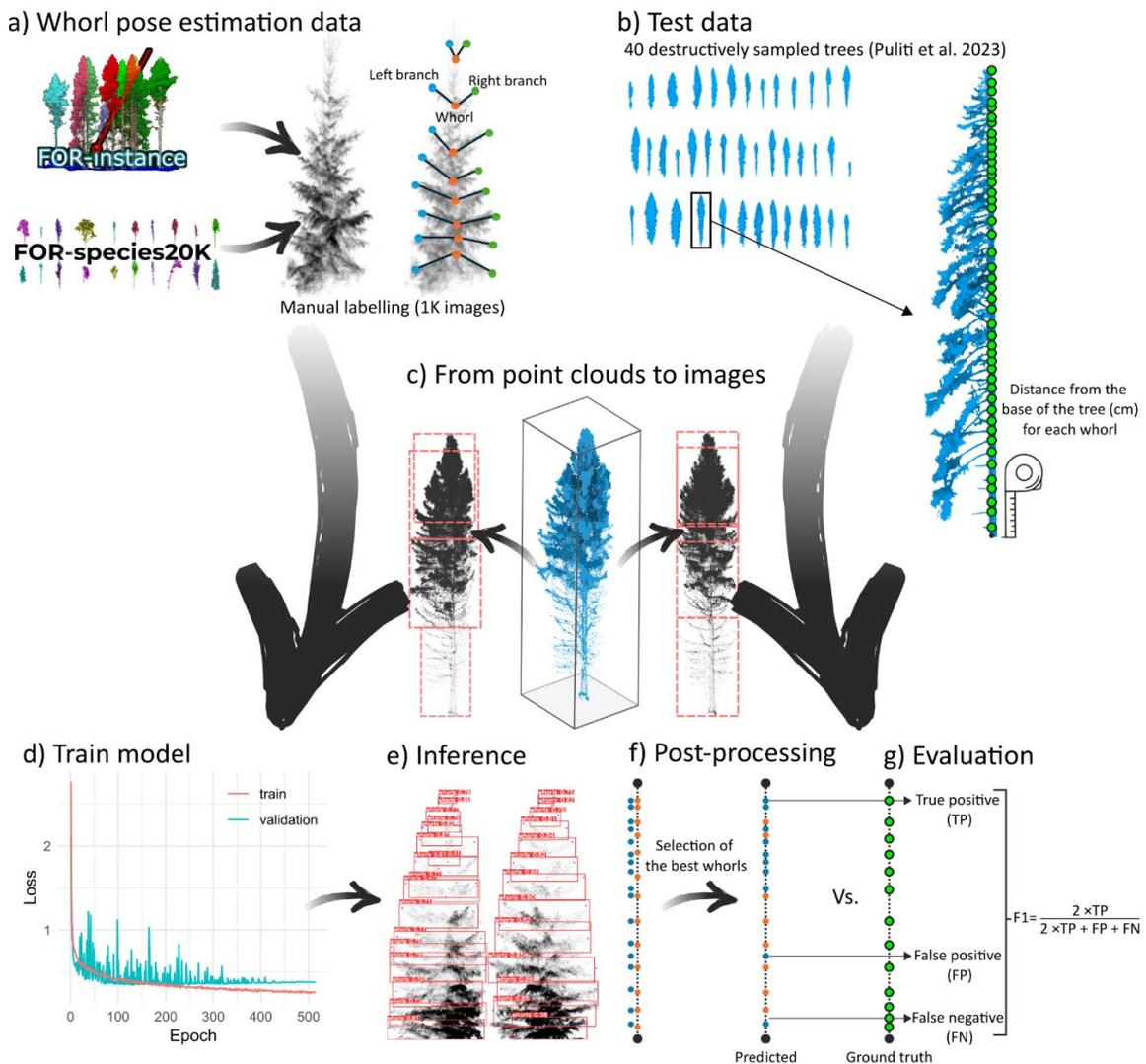

**Figure 1**. Visual workflow of the current study.

# 3. Methods

The following section describes the different steps adopted in the proposed method and Figure 1 shows the overall workflow. For inference, parallel processing is employed to handle multiple trees simultaneously, optimizing the workflow for large datasets.

## 3.1 From point cloud to image space

In the first step the developed method adopts the same approach as by Puliti et al. (2023a) where each individual tree point clouds are sliced (10 m wide slices) into radial sections. Each section is converted into black (with transparency) and white images, capturing the cross-sectional view of the tree at different heights (see figure 1c). To allow the detection of fine crown structures the images represented vertical tree sections of maximum 10 m in height, this resulting in tall trees (>10 m) being composed of several sections (see Figure 1c).

The functions to perform these steps are defined in the following functions: slice_tree_center_thick_slices, rotate_point_cloud, plot_section_as_image_with_alpha. These are then combined into convert_sections_to_images and then into process_point_cloud.

## 3.2 Whorl Detection Workflow

We selected the YOLOv8 framework for modeling whorl poses (Jocher et al. 2024). A nano model whorl detection model was trained for 515 epochs at an image size of 1000 pixels × 1000 pixels.

## 3.3 From image space to tree space

Post-processing of the whorl-pose detection is performed to:

- **Back-convert the predicted keypoints to real-world coordinates:** this is done by converting from yolo format coordinates (i.e. coordinates relative to the image width and height) to point cloud coordinates using metadata related to the image bounding box (minimum and maximum x and y for the image corners). See convert_to_real_world function for more implementation details.
- **Merge the ensemble of multi-view predictions**: This is done by merging all the z values for the whorl predictions onto a common z axis.
- **Filter the merged whorls**: by selecting the most confident prediction within each 25 cm of tree height (min_whorl_dist_m) we ensured that duplicate whorls are included, while at the same time allowing gap-filling for whorls detected only from one of the views.

Further based on the geometries of the selected whorls we computed branch insertion angles (degrees; see calculate_angle_at_p2 function) and maximum branch length (m; see calculate_distance function) calculated.

## 3.4 Benchmarking

The model was applied to the whole Puliti et al. (2023) test dataset using the pose_detection_tree function, which combines all of the functions described above.

To evaluate the accuracy of the whorl detection algorithm for each tree, the following metrics were calculated:

1. True Positives (TP): A detected whorl was classified as a true positive if it was within 20 cm (along the Z-axis) of a corresponding ground truth whorl measured in the field. This indicates that the whorl was correctly identified by the detection algorithm.
2. False Positives (FP or commission errors): A detected whorl was classified as a false positive if there was no corresponding ground truth whorl within 20 cm. This means that the detection algorithm incorrectly identified a whorl where none actually existed according to the field measurements.
3. False Negatives (FN or omission errors): A ground truth whorl was classified as a false negative if the detection algorithm failed to identify any corresponding whorl within 20 cm. This indicates that a whorl present in the field measurements was missed by the detection algorithm.

After identifying the TP, FP, and FN for each tree, the precision, recall, and F1-Score were calculated to evaluate the model's performance according to:

$$Precision = \frac{TP}{TP + FP} \qquad \text{Eq.1}$$

$$Recall = \frac{TP}{TP + FN} \qquad \text{Eq.2}$$

$$F - score = 2 \times \frac{P \times R}{P + R} \qquad \text{Eq.3}$$

# 3. Results and discussion

## 3.1 Evaluation

The proposed workflow successfully detected tree whorls in 67% of the trees in the dataset, omitted 30% of the whorls, and wrongly predicted 33% of the whorls that were not present in the ground truth data. This performance is comparable to that of Puliti et al. (2023) (refer to Table 1), who trained a whorl detection model using a small dataset of 141 labeled images from the same study area where the test trees were located. In contrast, the whorl-pose-detection model presented here was trained on a significantly larger dataset of 1,071 images, encompassing a broader range of laser scanning platforms (ULS, MLS, and TLS data) and tree species (P. abies, P. sylvestris) from outside the test study area. Therefore, it is encouraging to observe that our global model performs as well as a local model, demonstrating the transferability of the developed whorl-pose-detection model.

**Table 1**. Benchmarking results against results by Puliti et al. (2023)

|  | Precision | Recall | F1-score |
|---|---|---|---|
| **Puliti et al. (2023)** | **0.70** | 0.63 | 0.66 |
| **This study** | 0.69 | **0.68** | **0.68** |

Figure 2 illustrates the typical output, showing the detected whorls overlaid on the original point cloud. The method demonstrated high accuracy in identifying whorl positions and estimating branch angles, as confirmed by manual validation.

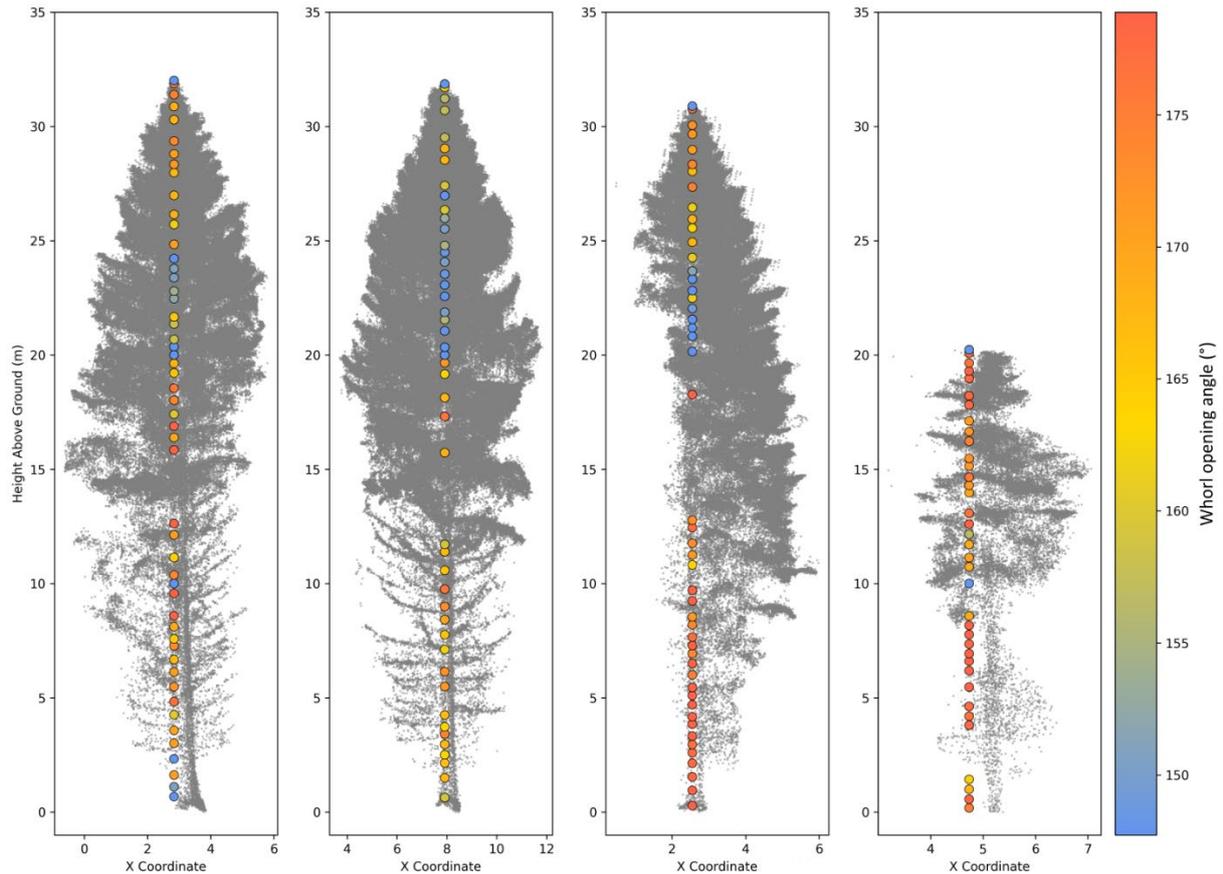

**Figure 2**. Typical output of the proposed method, i.e. detected whorl positions with associated branch opening angle (º) and maximum branch length (m).

The time required to process all trees with the proposed code is of approx. 5.7 seconds per tree on a Windows machine with 16 CPU cores and an NVIDIA Quadro P6000 GPU. This highlights the scalability potential of the method.

# 4. Conclusion

The presented automated workflow for tree whorl pose detection enables the efficient and robust capture of fine tree crown information for coniferous trees from laser scanning data and deep learning. The proposed workflow offers a scalable solution for forestry applications, enabling more detailed and accurate analysis of tree structures. The obtainable information is relevant for several downstream individual tree applications such as:

- **tree growth/site productivity**: As showcased by Puliti et al. (2023), information on tree growth can be derived from detected whorls and in turn this can be used to determine past and possibly future individual tree growth trajectories. Such information can provide a fine-scale mapping of site productivity, a key variable for forest management and carbon capture.
- **wood quality**: the presence of whorls corresponds to internal wood properties such as knots and the fibre disturbance of the wood around them, reducing structural timber properties. Further, information on the branch insertion angle gives information about the appearance of the knot within the produced products and the regarded wood quality.
- **competition**: The size of the crown, and with that the amount of visible whorls, is related to the competition for light. Supressed trees will develop smaller crowns with smaller branch diameters then dominant trees and can therefor give an understanding of the status of the respective tree within the stand.
- **traceability**: Whorl pattern along the stem and branch sizes vary from stem to stem and could be, in addition to other geometric features, used for a later recognition of the stems in the wood processing chain.

However, the accuracy of whorl detection can vary depending on the quality of the input data, suggesting that future work will explore the adaptation of this method to different forest environments and further optimization of the YOLO model for the different downstream forestry applications.

# Acknowledgments

This work is part of the Center for Research-based Innovation SmartForest: Bringing Industry 4.0 to the Norwegian forest sector (NFR SFI project no. 309671, smartforest.no).

# Data and Code Availability Statement

The data used in this study has been derived from publicly available FOR-instance (https://zenodo.org/records/8287792) and FOR-species20K (https://zenodo.org/records/13255198) datasets.

The code for applying our method to new data is available at:
https://github.com/stefp/whorl_pose_detector